\documentclass{article}
\usepackage{spconf,amsmath,graphicx}
\usepackage{soul,color}
\usepackage{algorithm}
\usepackage{algpseudocode}
\usepackage{amssymb}
\usepackage{enumitem}
\usepackage{cite}
\usepackage{tikz}
\usepackage{graphicx}
\newcommand\ChapterPrecis[2]{%
    \begin{tikzpicture}[remember picture,overlay]
    \node[anchor=north,draw,yshift=-#1] at (current page.north) {\parbox[t][1.6cm][c]{2.3\linewidth}{#2}};
    \end{tikzpicture}%
}

\title{Alzheimer's Disease Diagnostics by Adaptation of 
	3D Convolutional Network}
\name{Ehsan Hosseini-Asl\textsuperscript{1}\sthanks{Corresponding Author:---   Tel: (502) 852 3165, Fax: (502) 852 3940,  E-mail: ehsan.hosseiniasl@louisville.edu}, Robert Keynton\textsuperscript{2}, Ayman El-Baz\textsuperscript{2}}
\address{Author Affiliation(s)}

\address{\textsuperscript{1}Electrical and Computer Engineering Department, University of Louisville, Louisville, KY, USA.\\
\textsuperscript{2}Bioengineering Department, University of Louisville, Louisville, KY, USA.
}			

\begin{document}
\ninept
\maketitle
\begin{abstract}
Early diagnosis, playing an important role in preventing progress and treating the Alzheimer's disease (AD), is based on 
classification of features extracted from brain images. The features have to accurately capture main AD-related variations of 
anatomical brain structures, such as, e.g., ventricles size, hippocampus shape, cortical thickness, and brain volume. 
This paper proposed to predict the AD with a deep 3D convolutional neural network (3D-CNN), which can learn generic features 
capturing AD biomarkers and adapt to different domain datasets. The 3D-CNN is built upon a 3D convolutional autoencoder,
which is pre-trained to capture anatomical shape variations in structural brain MRI scans. Fully connected upper layers of
the 3D-CNN are then fine-tuned for each task-specific AD classification. Experiments on 
the \emph{CADDementia} MRI dataset with no skull-stripping preprocessing have shown our 3D-CNN outperforms 
several conventional classifiers by accuracy. Abilities of the 3D-CNN to generalize the  
features learnt and adapt  to other domains have been validated on the \emph{ADNI} dataset. 
\end{abstract}

\ChapterPrecis{1cm}{\textcopyright Copyright 2016 IEEE. Published in the IEEE 2016 International Conference on Image Processing (ICIP 2016), scheduled for September 25-28, 2016 in Phoenix, Arizona, USA. Personal use of this material is permitted. However, permission to reprint/republish this material for advertising or promotional purposes or for creating new collective works for resale or redistribution to servers or lists, or to reuse any copyrighted component of this work in other works, must be obtained from the IEEE. Contact: Manager, Copyrights and Permissions / IEEE Service Center / 445 Hoes Lane / P.O. Box 1331 / Piscataway, NJ 08855-1331, USA. Telephone: + Intl. 908-562-3966.}
\section{Introduction}
\label{sec:intro}

Alzheimer\textquoteright s disease (AD) is a progressive brain disorder and the most common case of dementia in the late life. AD leads to the death of nerve cells and tissue loss throughout the brain, thus reducing the brain volume in size dramatically through time and affecting most of its 
functions~\cite{mckhann1984clinical}. The estimated number of affected people will double for the next two decades, so that one out of 85 persons will have the AD by 2050~\cite{alzheimer20142014}. Because the cost of caring the AD patients is expected to rise dramatically, the necessity of having a computer-aided system for early and accurate AD diagnosis becomes critical~\cite{bron2015standardized}.  

Several popular non-invasive neuroimaging tools, such as structural MRI (sMRI), 
functional MRI (fMRI), and positron emission tomography (PET), have been investigated for developing such 
a system~\cite{jack2011introduction,mckhann2011diagnosis}. The latter
extracts features from the available images, and a classifier is trained to distinguish between different groups of 
subjects, e.g., AD, mild cognitive impairment (MCI), and normal control 
(NC) groups~\cite{bron2015standardized,cuingnet2011automatic,falahati2014multivariate,sabuncu2015clinical}. 
The sMRI has been recognized as a promising indicator of the AD progression~\cite{bron2015standardized,jack2013tracking}.

Various machine learning techniques were employed to leverage multi-view MRI, PET, and CSM data to predict 
the AD. Liu et al.~\cite{liu2015inherent} extracted multi-view features using several selected templates in the subjects' MRI dataset. 
Tissue density maps of each template were used then for clustering subjects within each class in order to extract an encoding 
feature of each subject. Finally, an ensemble of support vector machine (SVM) was used to 
classify the subject. Deep networks were also used for diagnozing the AD with different image modalities and clinical data. 
Suk et al.~\cite{suk2013deep} used a stacked autoencoder to separately extract features from MRI, PET, and cerebrospinal fluid (CSF) 
images; compared combinations of these features with due account of their clinical mini-mental state 
examination (MMSE) and AD assessment scale-cognitive (ADAS-cog) scores, and 
classified the AD on the basis of three selected MRI, PET, and CSF features with a multi-kernel SVM. Later on, a multimodal 
deep Boltzmann machine (BM) was used~\cite{suk2014hierarchical}  to extract one feature from each selected patch of 
the MRI and PET scans and predict the AD with an ensemble of SVMs. Liu et al.~\cite{liu2014multi} 
extracted 83 regions-of-interest (ROI) from the MRI and PET scans and used multimodal fusion to create a set of features to train 
stacked layers of denoising autoencoders. Li et al.~\cite{li2015robust} developed a multi-task deep learning for both AD classification and 
MMSE and ADAS-cog scoring by multimodal fusion of MRI and PET features into a deep restricted 
BM, which was pre-trained by leveraging the available MMSE and ADAS-cog scores.

\begin{figure*}[htb!]
	\scriptsize
	\begin{minipage}[b]{0.45\linewidth}
		\centering
		\centerline{\includegraphics[width=8cm]{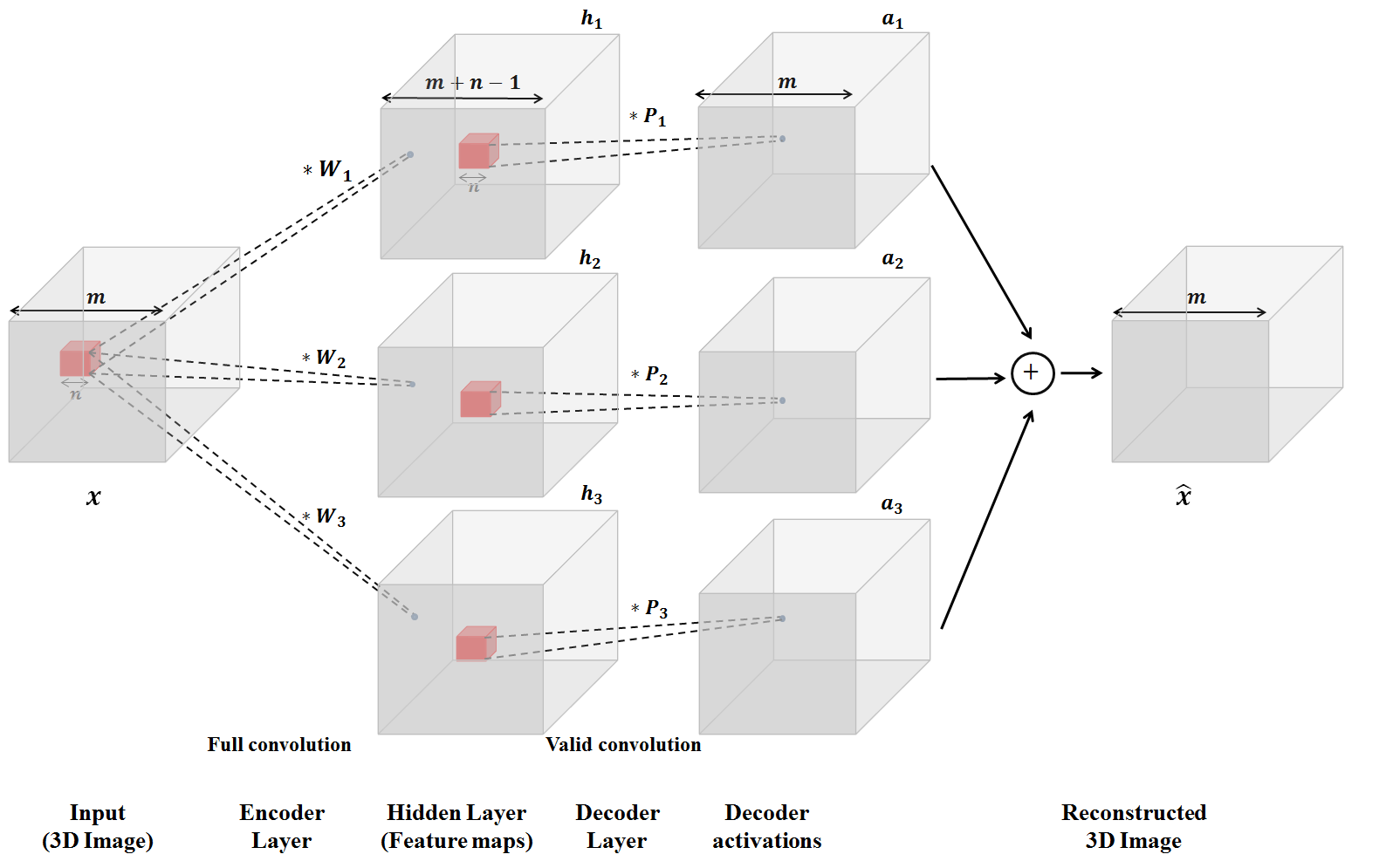}}
		{{\footnotesize (a)}}
	\end{minipage}
	\begin{minipage}[b]{0.55\linewidth}
		\centering
		\centerline{\includegraphics[width=8cm]{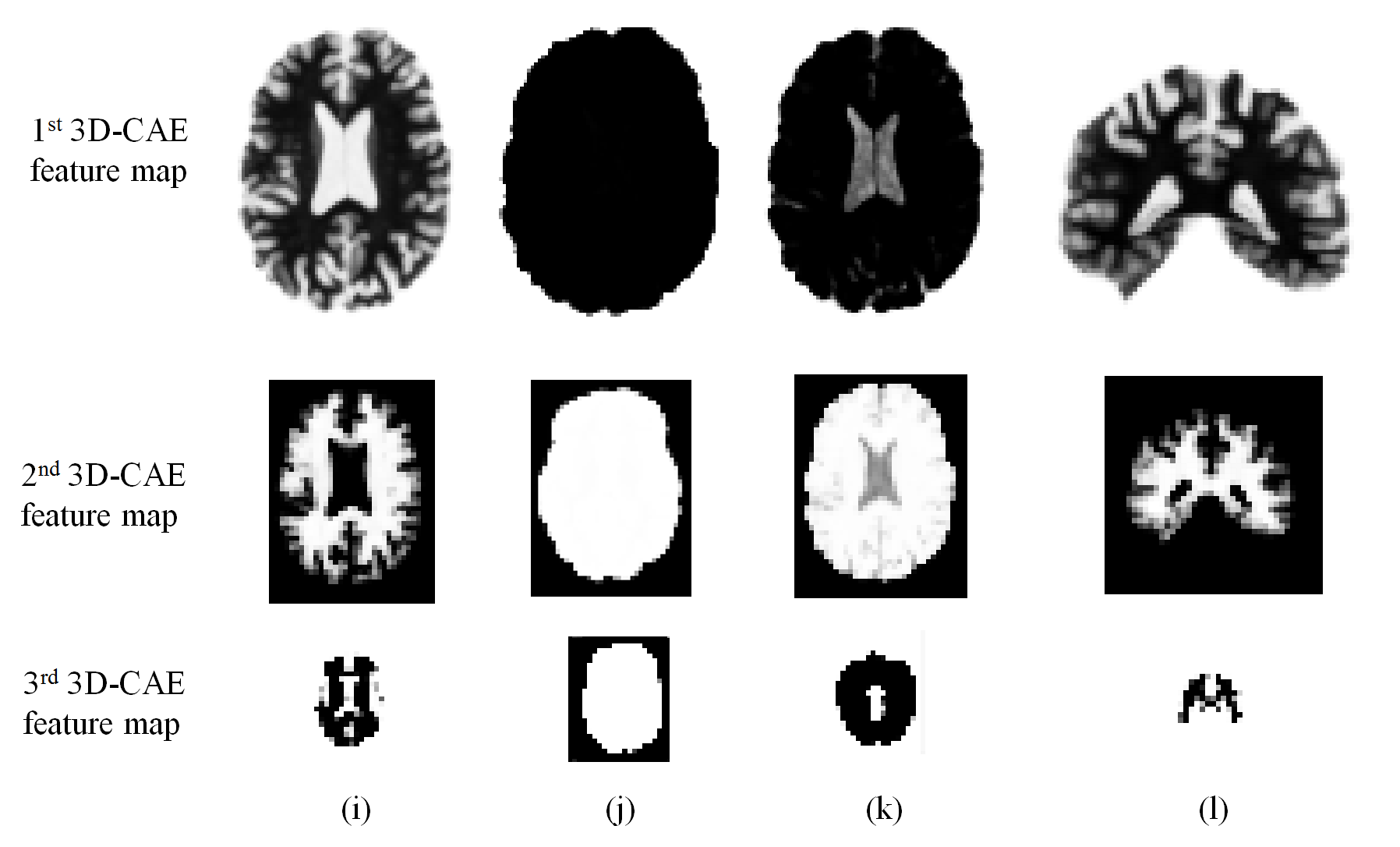}}
		{{\footnotesize (b)}}
	\end{minipage}
	\caption{(a) Schematic diagram of 3D Convolutional Autoencoder for feature extraction of a 3D Image based on reconstructing the input. Note that the image dimension increases in the encoding layer due to full convolution, and decreases to original dimension by valid convolution. (b) Selected slices of hierarchical 3D feature maps (trained on 30 subjects of~\emph{CADDementia}) in (i,j,k) axial and (l) sagittal view extracted at 3 layers of stacked 3D-CAE indicating (i) cortex thickness and volume, (j) brain size, (k) ventricle size, and (l) hippocampus model, extracted from the brain structural MRI. The feature maps are downsampled at each layer using max-pooling to reduce the size and detect higher level features.}\medskip
	\label{fig: 3D-CAE}
		\vspace{-6mm}
\end{figure*}

Voxel-wise, cortical thickness, and hippocampus shape-volume features of the sMRI are used to diagnose 
the AD~\cite{bron2015standardized}. The voxel-wise features are extracted after co-aligning (registering) all 
the brain image data to associate each brain voxel with a vector (signature) of multiple scalar measurements. 
Kl{\"o}ppel et al.~\cite{kloppel2008automatic} used the gray matter (GM) voxels as features and trained an SVM 
to discriminate between the AD and NC subjects. The brain volume in~\cite{fan2007compare} is segmented to 
GM, white matter (WM), and CSF parts, followed by calculating their voxel-wise densities and associating
each voxel with a vector of GM, WM, and CSF densities for classification.
For extracting cortical thickness features, Lerch et al.~\cite{lerch2008automated} segmented the registered brain 
MRI into the GM, WM, and CSF; fitted the GM and WM surfaces using deformable models; deform and expand the WM surface 
to the GM-CSF intersection; calculate distances between corresponding points at
the WM and GM surfaces to measure the cortical thickness, and use these features for classification. To quantify 
the hippocampus shape for feature extraction, Gerardin et al.~\cite{gerardin2009multidimensional} segmented 
and spatially aligned the hippocampus regions for various subjects and modeled their shape with a series of spherical harmonics. 
Coefficients of the series were then normalized to eliminate rotation--translation effects and used as features for 
training an SVM based classifier.

Comparative evaluations~\cite{bron2015standardized,cuingnet2011automatic,falahati2014multivariate,sabuncu2015clinical}
revealed several limitations of the above feature extraction techniques for classifying the AD. The voxel-wise feature vectors 
obtained from the brain sMRI are very noisy and can be used for classification only after smoothing and clustering to reduce 
their dimensionality~\cite{fan2007compare}. The cortical thickness and hippocampus model features neglect correlated shape 
variations of the whole brain structure affected by the AD in other ROIs, e.g., the ventricle's volume. Appropriateness of the 
extracted feature vectors highly depend on image preprocessing due to registration errors and noise, so that feature engineering 
requires the domain expert knowledge. Most of the trainable classifiers are biased toward a particular dataset, which was used for 
training and testing (i.e., the classification features extracted at the learning stage are dataset-specific). 

We propose a new deep 3D convolutional neural network (3D-CNN), for unsupervised generic and transferable feature extraction based on 3D extension of convolutional autoencoder (3D-CAE)~\cite{masci2011stacked} to overcome the aforementioned limitations in feature extraction from brain sMRI for AD classification. The proposed network combines a pretrained 3D-CAE in the source domain, e.g. \emph{CADDementia}, with upper task-specific layers to be fine-tuned in the target 
domain, e.g. \emph{ADNI} dataset,~\cite{yosinski2014transferable,long2015learning}. Such adaptation of pre-learned generic features allows for
calling the proposed classifier a  3D Adaptable CNN (3D-ACNN).

\vspace{-3mm}
\section{Model}
\label{sec:method}

The proposed AD diagnostic framework extracts features of a brain MRI with a source-domain-trained 3D-CAE and performs 
task-specific classification with a target-domain-adaptable 3D-CNN. The 3D-CAE architecture and the AD diagnosis 
framework using the 3D-ACNN are detailed in Sections~\ref{sec:3DCAE} and~\ref{sec:3DCNN},
respectively.

  \begin{figure*}[htb!]
  	\centering
  	\includegraphics[width=17cm]{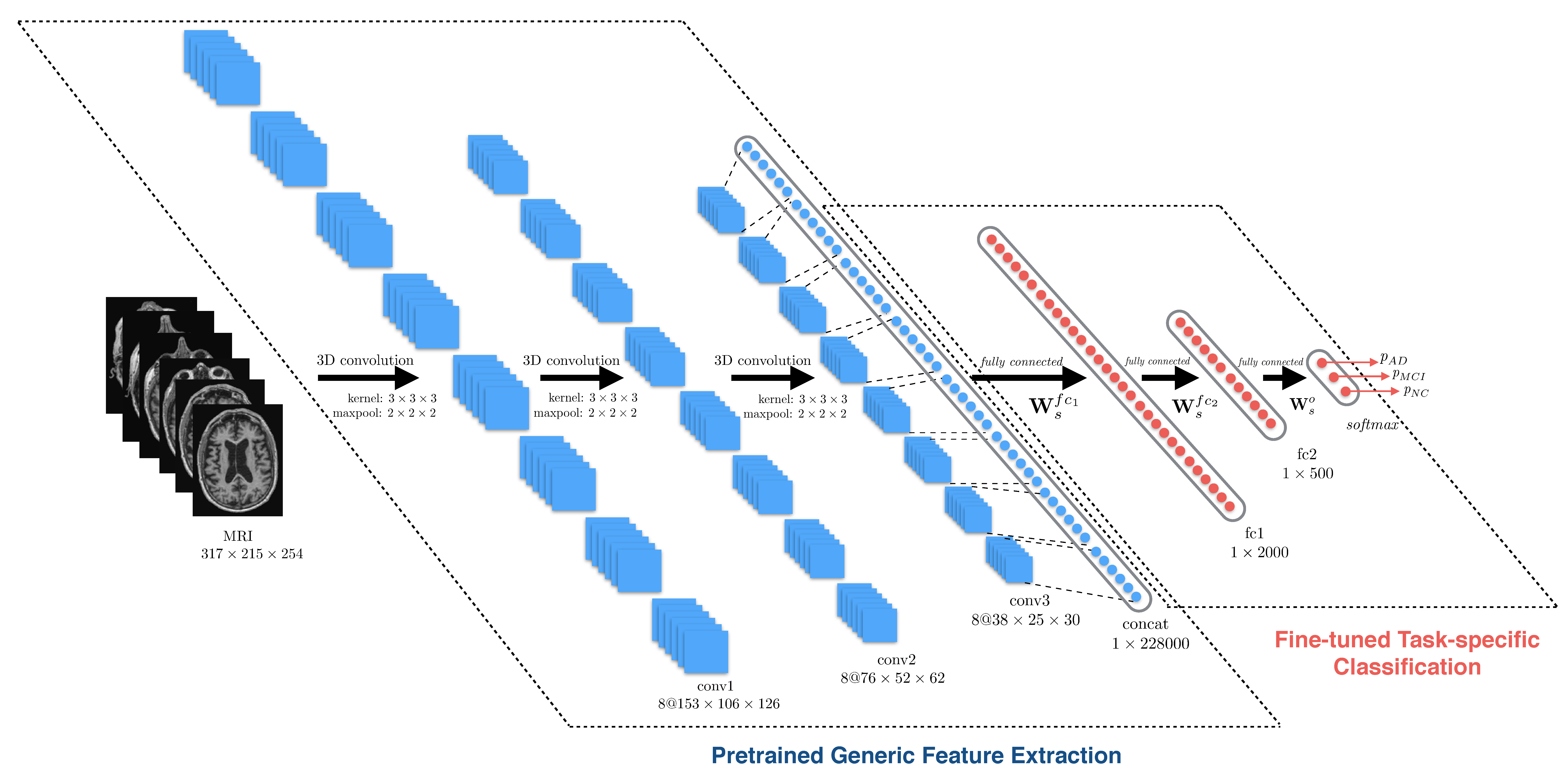}
  	\caption{Our 3D adaptable CNN (3D-ACNN) for Alzheimer's disease diagnosis.}
  	\label{fig_model}
  	\vspace{-3mm}
  \end{figure*}
  
\vspace{-3mm}
\subsection{Feature Extraction using 3D Convolutional Autoencoder:}
\label{sec:3DCAE}
Conventional unsupervised autoencoder extracts a few co-aligned scalar feature maps for a set of input 3D images 
with scalar or vectorial voxel-wise signals by combining data encoding and decoding. 
The input image is encoded by mapping each fixed voxel neighborhood to a vectorial feature space in the hidden layer and 
is reconstructed back in the output layer to the original image space. To extract features that capture
characteristic patterns of input data variations, training of the autoencoder employs back-propagation and constraints on 
properties of the feature space to reduce the reconstruction error. 

Extracting global features from 3D images with vectorial voxel-wise signals is computationally expensive and requires too 
large training data sets, due to growing fast numbers of parameters to be evaluated in the input (encoding) 
and output (decoding) layers~\cite{lecun1998gradient}. Moreover, local features are more suitable for extracting patterns from 
high-dimensional images, although autoencoders with full connections between all nodes of the layers try to 
learn global features. To overcome this problem, we use a stack of unsupervised CAE with locally 
connected nodes and shared convolutional weights to extract local features from 3D images with possibly long 
voxel-wise signal vectors~\cite{masci2011stacked,makhzani2014winner,Leng20153d}. Each input image is reduced 
hierarchically, using the hidden feature (activation) map of each CAE for training the next-layer CAE.

Our 3D extension of a hierarchical CAE proposed in~\cite{masci2011stacked} is sketched in Fig.~\ref{fig: 3D-CAE}(a). 
To capture characteristic variations of an input 3D image, $\mathbf{x}$, each voxel-wise feature, $h_{i:j:k}$, associated with the 
$i$-th 3D lattice node, $j$-th component of the input voxel-wise signal vector, and $k$-th feature map; $k=[1,\ldots,K]$, 
is extracted by a moving-window convolution 
(denoted below $\ast$) of a fixed $n\times n \times n$ neighborhood, $\mathbf{x}_{i:\mathrm{neib}}$,
of this node with a linear encoding filter, specified by its weights, $\mathbf{W}_k=[\mathbf{W}_{j:k}:\:j=1,\ldots,J]$ for each relative
neighboring location with respect to the node $i$ and each voxel-wise signal component $j$, 
followed by feature-specific biases, $\mathbf{b}_k=[b_{j,k}:\:j=1,\ldots,J]$ 
and non-linear transformations with a certain activation function, $f(\cdot)$: 
\begin{equation}\label{eq:encode}
h_{i:j:k} = f\left(\mathbf{W}_k\ast\mathbf{x}_{i:\mathrm{neib}} + b_{j:k}\right)
\end{equation}
The latter function is selected from a rich set of  constraining differentiable one, including, in particular, the 
sigmoid, $f(u) = (1+\exp(-u))^{-1}$ and rectified linear unit (ReLU), $f(u)=\max{(0,u)}$~\cite{glorot2010understanding}. Since the
3D image $\mathbf{x}$ in Eq.~(\ref{eq:encode}) has the $J$-vectorial voxel-wise signals, actually, the weights $\mathbf{W}_k$ define
a 3D moving-window filter convolving the union of $J$-dimensional signal spaces for each voxel within the window. 

To simplify notation, let $\mathbf{h}_k=\mathbb{T}\left(\mathbf{x}:\:\mathbf{W}_k,\mathbf{b}_k,f(\cdot)\right)$ 
denote the entire encoding of 
the input 3D image with $J$-vectorial voxel-wise signals with the $k$-th 3D feature map, $\mathbf{h}_k$, such that its 
scalar components are obtained with Eq.~(\ref{eq:encode}) using the weights $\mathbf{W}_k$ and 
bias vectors $\mathbf{b}_k$ for a given voxel neighborhood. 
The like inverse transformation, $\mathbf{T}_{\mathrm{inv}}(\ldots)$, with the same voxel neighborhood, but 
generally with the different convolutional weights, $\mathbf{P}_k$,  biases, $\mathbf{b}_{\mathrm{inv}:k}$, and, 
possibly, activation function, $g(\cdot)$,
decodes, or reconstructs the initial 3D image:
\begin{equation}\label{eq:decode}
\widehat{\mathbf{x}} = \sum\limits_{k=1}^K 
\underbrace{\mathbf{T}_{\mathrm{inv}}\left(\mathbf{h}_k:\:\mathbf{P}_k,\mathbf{b}_{\mathrm{inv}:k}, g(\cdot)\right)}_{\mathbf{a}_k}
\end{equation}
Given $L$ encoding layers, each layer $l$ generates an output feature image, $\mathbf{h}_{(l)} = [\mathbf{h}_{(l):k}:\:k=1,\ldots,K_l]$,
with $K_l$-vectorial voxel-wise features and receives the preceding output, $\mathbf{h}_{(l-1)} = [\mathbf{h}_{(l-1):k}:\:k=1,\ldots,K_{l-1}]$ 
as the input image (i.e., $\mathbf{h}_{(0)} = \mathbf{x}$.

The 3D-CAE of Eqs.~(\ref{eq:encode}) and~(\ref{eq:decode}) is trained by minimizing the mean squared reconstruction error 
for $T$; $T\ge1$, given training input images,
$\mathbf{x}^{[t]}$; $t=1,\ldots,T$, 
\begin{equation}
E(\boldsymbol{\theta})=\frac{1}{T}\sum_{t=1}^{T}\parallel\widehat{\mathbf{x}}^{[t]}- \mathbf{x}^{[t]}\parallel^{2}_{2}
\label{eq: cost}
\end{equation}
where $\boldsymbol{\theta}=[\mathbf{W}_{k};\mathbf{P}_k;\mathbf{b}_{k};\mathbf{b}_{\mathrm{inv}:k}:\:k=1,\ldots,K]$,
and $\parallel\ldots\parallel^{2}_{2}$ denote all free parameters and
the average vectorial $\ell_{2}$-norm over the $T$ training images, respectively. To reduce the number of 
the free parameters, the decoding, $\mathbf{P}_{k}$, and encoding, $\mathbf{W}_{k}$, weights were tied by flipping 
over all their dimensions as proposed in~\cite{masci2011stacked}. The cost of Eq.~(\ref{eq: cost}) was minimized in the parameter
space by using the stochastic gradient descent search, combined with error back-propagation.

\subsection{AD Classification by 3D Adaptive CNN (3D-ACNN)}
\label{sec:3DCNN}
While the lower layers of a goal predictive 3D-CNN extract generalized features, the upper layers have to facilitate task-specific 
classification using those features~\cite{long2015learning}. The proposed classifier extracts the generalized features by using
a stack of locally connected bottom convolutional layers, while performing task-specific fine-tuning of parameters of the fully connected 
upper layers. Training the proposed hierarchical 3D-CNN consists of pre-training, initial training of the lower convolutional layers,
and final task-specific fine-tuning. At the pre-training stage, the convolutional layers for generic feature extraction are formed as a stack
of 3D-CAEs, which were pre-trained in the source domain. Then these layers are initialized by encoding the 3D-CAE 
weights~\cite{chen2015net2net}, and, finally, the deep-supervision-based~\cite{lee2014deeply} fine-tuning of the upper 
fully-connected layers is performed for each task-specific binary or multi-class sMRI classification.

Due to pre-training on the source domain data, the bottom convolutional layers can extract generic features related to the AD 
biomarkers, such as the ventricular size, hippocampus shape, and cortical thickness, as shown in Fig.~\ref{fig: 3D-CAE}(b). 
We use the Net2Net initialization~\cite{chen2015net2net}, which allows for different convolutional kernel and pool sizes of the 
3D-CNN layers, comparing to those in the 3D-CAE, based on the target-domain image size and imaging specifications, and 
facilitates adapting the 3D-CNN across the different domains. To classify the extracted features in a task-specific way,
weights of the upper fully-connected 3D-CNN  layers are fine-tuned on the target-domain data by minimizing 
a specific loss function. The loss depends explicitly on the weights and is proportional to a negated log-likelihood of the 
true output classes, given the input features extracted from the training target-domain images by the 
pre-trained bottom part of the network. 

Our implementation of the 3D-CNN uses the ReLU activation functions at each inner layer and
the fully connected upper layers with a softmax top-most output layer (Fig.~\ref{fig_model}), predicting the probability of belonging an input brain sMRI to the AD, MCI, or NC group. The Adadelta gradient descent~\cite{zeiler2012adadelta} was used to 
update the pre-trained 3D-CAE and fine-tune the entire 3D-ACNN.

\vspace{-3mm}
 \begin{table}[htb!]
 	\caption{Demographic data for 210 subjects from the ADNI database (STD -- standard deviation).}
 	\centering
 	\begin{tabular}{lccc}
 		\hline
 		Diagnosis & AD & MCI & NC \\
 		\hline
 		Number of subjects & 70 & 70 & 70 \\
 		Male / Female & 36 / 34 & 50 / 20 & 37 / 33 \\
 		Age (mean$_{\pm\mathrm{STD}}$) & $75.0_{\pm 7.9}$ & $75.9_{\pm 7.7}$ & $74.6_{\pm 6.1}$ \\
 		\hline
 	\end{tabular}		
 	\label{tab:demographic}
 \end{table}
 
 \begin{table*}[htb!]
 	\caption{Comparative performance (ACC,\%) of our classifier vs. seven competitors (n/a -- non-available).}
 	\centering
\resizebox{0.85\textwidth}{!}{
 	\begin{tabular}{llccccc}
 		\hline
 		{} & {} &\multicolumn{5}{c}{ Task-specific classification [$\mathrm{mean}_{\mathrm{STD}}$,\%].}\\
 		\cline{3-7}
 		Approach & Modalities & AD/MCI/NC & AD+MCI/NC & AD/NC & AD/MCI & MCI/NC \\
 		\hline
 		Suk et al.~\cite{suk2013deep}           & PET+MRI+CSF &  $\textrm{n/a}$ & $\textrm{n/a}$ & $95.9_{1.1}$ &  $\textrm{n/a}$ & $85.0_{1.2}$ \\
 		Suk et al.~\cite{suk2014hierarchical}          & PET+MRI &  $\textrm{n/a}$ & $\textrm{n/a}$ & $95.4_{5.2}$ &  $\textrm{n/a}$ & $85.7_{5.2}$ \\
 		Zhu et al.~\cite{zhu2014novel}          & PET+MRI+CSF &   $\textrm{n/a}$ & $\textrm{n/a}$ & $95.9_{\textrm{n/a}}$ & $\textrm{n/a}$ & $82.0_{\textrm{n/a}}$ \\
 		Zu et al.~\cite{zu2015label}                         & PET+MRI &   $\textrm{n/a}$ & $\textrm{n/a}$ & $96.0_{\textrm{n/a}}$ & $\textrm{n/a}$ & $80.3_{\textrm{n/a}}$ \\
 		Liu et al.~\cite{liu2014multi}                         & PET+MRI & $53.8_{4.8}$ & $\textrm{n/a}$ & $91.4_{5.6}$ & $\textrm{n/a}$ & $82.1_{4.9}$\\
 		Liu et al.~\cite{liu2015inherent}                             & MRI & $\textrm{n/a}$& $\textrm{n/a}$ & $93.8_{\textrm{n/a}}$ & $\textrm{n/a}$ & $89.1_{\textrm{n/a}}$\\
 		Li et al.~\cite{li2015robust}                   & PET+MRI+CSF & $\textrm{n/a}$&  $\textrm{n/a}$ & $91.4_{1.8}$ & $70.1_{2.3}$ & $77.4_{1.7}$\\
 		Sarraf et al.~\cite{sarraf2016classification}   & fMRI & $\textrm{n/a}$&  $\textrm{n/a}$ & $96.8_{\textrm{n/a}}$ & $\textrm{n/a}$ & $\textrm{n/a}$\\
 		Our 3D-ACNN 
 		& MRI & $\mathbf{89.1_{1.7}}$ & $\mathbf{90.3_{1.4}}$ & $\mathbf{97.6_{0.6}}$ & $\mathbf{95_{1.8}}$ & $\mathbf{90.8_{1.1}}$ \\
 		\hline
 	\end{tabular}
 }
 	\label{tab:comp}
 \end{table*}
  
 \section{Experimental Results}
 To performance of the proposed 3D-CAES is evaluated on \emph{CADDementia}~\footnote{http://caddementia.grand-challenge.org.} as source domain, for generalized feature extraction. The data set contains structural T1-weighted MRI (T1w) scans of patients with the diagnosis of probable AD, patients with the diagnosis of MCI, and NC without a dementia syndrome~\cite{bron2015standardized}. To pretrain 3D-CAES on \emph{CADDementia}, sMRI are preproceesed by spatially normalizing using rigid registration approach. Then skull is removed and image intensities are normalized to $[0,1]$, resulting in sMRI of size $(200\times 150\times 150)$. 
   
   The classification performance of proposed 3D-ACNN is evaluated on AlzheimerÕs Disease Neuroimaging Initiative (ADNI) database, as target domain, for five classification tasks: four binary ones 
   (AD vs. NC, AD+MCI vs. NC, AD vs. MCI, MCI vs NC) and the ternary classification (AD vs. MCI vs. NC). Classification accuracy was evaluated for each test by ten-fold cross-validation. ADNI dataset without using any preprocessing and skull stripping is used AD classification, compared to preprocessed \emph{CADDementia} dataset. The Theano library~\cite{Bastien-Theano-2012}
 was used to develop the deep CNN implemented for our experiments on the Amazon EC2 g2.8xlarge instances 
 with GPU GRID K520.

 To pretrain 3D-CAE for feature extraction of brain sMRI, we use ReLu nonlinear function as the encoder and decoder layer's activation. The 3D-CAE contains eight encoding and decoding filters of size $(3\times 3\times 3)$. Three hierarchical 3D-CAE's are trained to extract the low-dimensional feature maps. The extracted feature maps are of dimension $(102\times 76\times 76)$, $(52\times 40\times 40)$, and $(28\times 22\times 22)$, subsequently. Each 3D-CAE extracts eight feature maps, according to the number of their encoding filters (Fig.~\ref{fig: 3D-CAE}(b)). Selected slices of the three feature maps from 
 each layer of our stacked 3D-CAE (abbreviated 3D-CAES below) in Fig.~\ref{fig: 3D-CAE}(b), show that the learnt generic convolutional 
 filters can really capture features related to the AD biomarkers, e.g., the ventricle size, cortex thickness, and hippocampus model. 
 These feature maps were generated by the pre-trained 3D-CAES for the CADementia database. According to these projections, the 
 first layer of the 3D-CAES extracts the cortex thickness as a discriminative AD feature of AD, whereas 
 the brain size (related to the patient gender), size of ventricles, and hippocampus model are represented by the subsequent layers. 
 Each 3D-CAES layer combines the extracted lower-layer feature maps in order to train the higher level for describing more in detail 
 the anatomical variations of the brain sMRI . Both the ventricle size and cortex thickness features are combined to extract 
 conceptually higher-level features at the next layers.

Performance of the proposed 3D-ACNN classifier, in terms of accuracy (ACC), for each task-specific classification was evaluated 
and compared to competing approaches~\cite{suk2013deep,suk2014hierarchical,zhu2014novel,liu2014multi,liu2015inherent,li2015robust,sarraf2016classification,zu2015label}. Table~\ref{tab:comp} presents the average results of ten-fold cross-validation of our classifier. According to these experiments, the proposed 3D-ACNN outperforms the other approaches in all five task-specific cases, in spite of employing only a single imaging
modality (sMRI) and performing no prior skull-stripping.

  \section{Conclusion and Future Work}
  \vspace{-3mm}
  This paper proposed a 3D-ACNN classifier, which can more accurately predict the AD on structural brain MRI scans, than 
  several other state-of-the-art predictors. The pretraining and layer freezing were used to enhances generality of features in capturing the AD biomarkers. Three stacked 3D CAE network were pretrained on \emph{CADDementia} Dataset. Then the learnt features are extracted and used as AD biomarkers detection in bottom layers of a 3D CNN network. Then three fully connected layers are stacked on top of the bottom layers to perform AD classification on 210 subjects of ADNI dataset. Classification performance were measured using ten-fold crossvalidation, and were compared to the state-of-the-art models, demonstrated the out-performance of the proposed 3D CNN. 
  The future application of proposed 3D-ACNN include detection of lung cancer~\cite{el2003unified,el2009automatic,el20113d,farag2006appearance}, heart failure~\cite{khalifa2012accurate}, and autism detection~\cite{casanova2011quantitative}.
  
\bibliographystyle{IEEEbib}
\bibliography{references_abv}

\end{document}